\documentclass[a4paper,fleqn]{cas-dc}

\usepackage[numbers]{natbib}

\def\tsc#1{\csdef{#1}{\textsc{\lowercase{#1}}\xspace}}
\tsc{WGM}
\tsc{QE}

\usepackage{caption}

\newcommand{\at}{44.0}
\newcommand{\aT}{73.4}
\newcommand{\bt}{44.5}
\newcommand{\bT}{73.8}

\newcommand{\ct}{46.8}
\newcommand{\cT}{76.0}

\newcommand{\CMEMsomet}{49.1}
\newcommand{\CMEMsomett}{52.4}
\newcommand{\CMEMsomeT}{78.2}
\newcommand{\CMEMsomeTT}{80.9}

\newcommand{\CLIMsomet}{47.6}

\newcommand{\CLIMsomeT}{76.5}

\newcommand{\CMEMCLIMsomet}{49.7}
\newcommand{\CMEMCLIMsomett}{52.9}

\newcommand{\CMEMCLIMsomeT}{78.4}
\newcommand{\CMEMCLIMsomeTT}{81.4}

\newcommand{\CMEMCLIMucft}{95.3}

\newcommand{\CMEMCLIMhmdbt}{74.5}

\begin{document}
\let\WriteBookmarks\relax
\def\floatpagepagefraction{1}
\def\textpagefraction{.001}
\let\printorcid\relax       

\shorttitle{Behavior Recognition Based on the Integration of Multigranular Motion Features}

\shortauthors{L. Zhang}

\title [mode = title]{Behavior Recognition Based on the Integration of Multigranular Motion Features}

\author[Address1]{Lizong Zhang}
\ead{l.zhang@uestc.edu.cn}

\author[Address1]{Yiming Wang}
\ead{yumy@std.uestc.edu.cn}

\author[Address2]{Bei Hui}
\cormark[1]
\cortext[cor1]{Corresponding author}
\ead{bhui@uestc.edu.cn}

\author[Address3]{Xiujian Zhang}
\ead{zxjwell@163.com}

\author[Address4]{Sijuan Liu}
\ead{liusj@swufe.edu.cn}

\author[Address1]{Shuxin Feng}
\ead{fsxscholar@163.com}

\address[Address1]{School of Computer Science and Engineering, University of Electronic Science and Technology of China, Chengdu 611731, China}
\address[Address2]{School of Information and Software Engineering, University of Electronic Science and Technology of China, Chengdu 611731, China.}
\address[Address3]{Key Laboratory of Artificial Intelligence Measurement and Standards for State Market Regulation, Beijing Aerospace Institute for Metrology and Measurement Technology, Beijing 100076, China}
\address[Address4]{Research Institute of Social Development, Chengdu 610074, Southwestern University of Finance and Economics}

\begin{abstract}
    The recognition of behaviors in videos usually requires a combinatorial analysis of the spatial information about objects and their dynamic action information in the temporal dimension. Specifically, behavior recognition may even rely more on the modeling of temporal information containing short-range and long-range motions; this contrasts with computer vision tasks involving images that focus on the understanding of spatial information. However, current solutions fail to jointly and comprehensively analyze short-range motion between adjacent frames and long-range temporal aggregations at large scales in videos. In this paper, we propose a novel behavior recognition method based on the integration of multigranular (IMG) motion features. In particular, we achieve reliable motion information modeling through the synergy of a channel attention-based short-term motion feature enhancement module (CMEM) and a cascaded long-term motion feature integration module (CLIM). We evaluate our model on several action recognition benchmarks such as HMDB51, Something-Something and UCF101. The experimental results demonstrate that our approach outperforms the previous state-of-the-art methods, which confirms its effectiveness and efficiency.

\end{abstract}


\ExplSyntaxOn
\keys_set:nn { stm / mktitle } { nologo }
\ExplSyntaxOff
\maketitle


\section{Introduction}

Recent years have witnessed a tremendous growth in the amount of video data on the Internet, resulting in the need for intelligent and autonomous video analysis technology \cite{cai2018private, cai2019trading}. 
Behavior recognition, as a fundamental task of video analysis technology, has become increasingly demanding in video-based applications such as human–machine interaction, autonomous driving, and intelligent surveillance \cite{andersson2019deep, ziaeefard2015semantic}. 
However, recognizing the motion information of objects is nontrivial due to occlusion, dynamic backgrounds, and moving cameras in video scenarios \cite{li2021occlusion, cai2016collective}. 
For example, it is difficult to distinguish between behaviors when faced with dynamic and moving backgrounds.
Therefore, this paper introduces a novel model for capturing temporal structure and spatial features to model motion features.

To comprehensively apply temporal features, current recognition methods mainly focus on two types of temporal motion information: short-range and long-range motions.
Specifically, short-range motion refers to the changes between similar adjacent frames at the pixel level. 
This type of motion may provides instant motion information. 
The issue is how to distinguish and represent the differences among these frames. 
Alternatively, long-range motion refers to the semantic associations between distant frames in videos. 
It can integrate short-range motion features to learn the interaction information among them. 
The challenge is to search for correlations among distant frames.

For modeling short-range motion features, current methods \cite{simonyan2014two} mainly apply optical flows to directly represent the temporal motion features of short distances, 
then fuse them with the spatial features extracted from RGB frames. 
These approaches can distinguish the differences between adjacent frames due to the frame gradients represented in the optical flows.
However, the corresponding training processes of optical flows and RGB streams are usually separated, where information fusion occurs only at the end of these approaches. 
Thus, such two-stream architectures result in no effective interaction process between spatial and temporal motion features.

For the modeling of long-range motion features, two types of solutions are available. 
The first category of methods, such as temporal segment networks (TSN) \cite{wang2016temporal}, directly apply the average pooling and multiscale temporal window integration to represent the long-range features of videos. 
Another category applies 3D or (2+1) D convolutional neural networks (CNN) \cite{ji20123d, tran2015learning} to implement joint spatial and temporal motion feature modeling. 
These methods model the correlations among distant frames and learn long-range features through many local convolution operations. 
However, both types of solutions are still insufficient for modeling detailed long-term temporal information.
The first type of solution may easily lose long-term temporal information due to its simple integration strategy.
The second type requires a massive volume of repetitive local convolution operations, leading to optimization difficulty \cite{he2016deep} as long-range motion features can hardly propagate through a long path. 
Thus, it is still a challenging task to capture the effective spatiotemporal information of long-range motion features.

\begin{figure}[ht]  
\centering
\includegraphics[width = \linewidth]{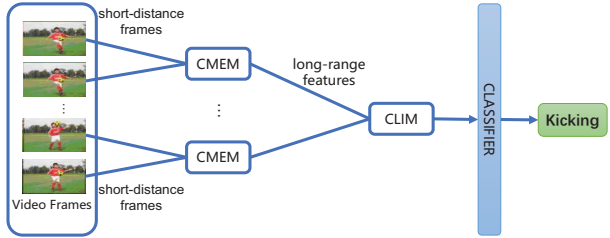}
\caption{Overview of our proposed modules. The CMEM is proposed to expand the differences between short-distance motion features, while the CLIM is proposed to enhance the probability of interaction between long-range features.}
\label{fig1}
\end{figure}

To address the above issues, this paper proposes a behavior recognition approach called the integration of multigranularity (IMG), which integrates two different modules for modeling two types of temporal and spatial motion information. 
As shown in Fig.~\ref{fig1}, IMG is composed of a channel attention-based motion feature enhancement module (CMEM) for short-distance motions between adjacent frames,
and a cascaded long-term motion feature integration module (CLIM) for further long-range temporal aggregation at large scales.

Specifically, the CMEM excites the motion-sensitive part of the original features and avoids polarization through a joint multiscale learning process involving spatial and temporal motion features. 
Concretely, residual connection \cite{he2016deep} and channel attention techniques \cite{hu2018squeeze} are applied to enhance the feature-level interactions between spatial and temporal motion information. 
In this way, the CMEM can realize effective modeling among short-range spatial and temporal motion features. 
Moreover, the CLIM realizes information interaction among the extracted motion features under different receptive fields through a cascaded residual network structure,
which is equivalent to repeated long-range spatiotemporal interactions, 
and the residual network structure also prevents the degradation of the network. 
Then combined with an adaptive shift module, which increases the interaction frequencies of long-range frames, 
The CLIM can achieve long-range temporal aggregation at large scales. 
Finally, two modules are combined with a residual network.
Through extensive experiments, our model is able to significantly improve the expression of short-range and long-range motion features, 
effectively enhance the behavior recognition accuracy of the final model, 
and achieve state-of-the-art results on several datasets.
To summarize, the contributions of our work include the following.

\begin{itemize}
\item We provide a novel behavior recognition method that integrates the two modules listed below to reliably model motion features. 
\item The CMEM addresses the poor timeliness and insufficient interaction between spatial features and temporal motion features exhibited by existing methods.
\item The CLIM enhances the interaction frequencies of long-range motion features with an adaptive shift module.
\item Extensive evaluation results are used to demonstrate the advancements of the proposed model.
\end{itemize}

The rest of this paper is organized as follows: chapter 2 briefly introduces the existing behavior recognition methods. Chapter 3 describes our proposed behavior recognition method in detail. Chapter 4 evaluates the effectiveness of our proposed method through experiments. Chapter 5 is the summary of the paper.
\section{Related Work}

In the current field of video behavior recognition, deep learning has gradually become the mainstream approach \cite{zheng2020privacy, cai2021generative}. 
For these methods, effectively modeling the motion information of videos is the key to their performance. 
The existing methods usually focus on two types of motion information: 1) short-range motion between adjacent frames and 2) long-range temporal aggregation at large scales.

For modeling the short-range motions between adjacent frames, current methods are mainly implemented by optical flow extraction and two-stream architectures. 
The classical two-stream architecture \cite{simonyan2014two} consists of a 2D CNN that learns spatial features from RGB frames and a temporal 2D CNN that models temporal motion information in the form of optical flows. 
Then, many successive works extended this model. 
Different combination strategies for fusing RGB streams and optical flows have been proposed. 
For example, Feichtenhofer \cite{feichtenhofer2016convolutional, feichtenhofer2017spatiotemporal} proposed extracting the features of the temporal dimension by using Conv3D and Pool3D, and fusing the features in the temporal dimension. 
After that, a sparse sampling strategy that is used to capture feature associations between frames was proposed through a TSN \cite{wang2016temporal}. 
However, the methods mentioned above have two common shortcomings. 
One is that these methods require additional computation and storage costs to deal with optical flows. 
The second is that the training processes for RGB streams and optical flows are separated, which leads to extremely limited interactions between these two types of features. 
Fusion is performed only in the late layers, which makes it difficult to generate effective feature-level interactions. 

To solve the problems described above, some works, such as STM \cite{jiang2019stm} have tried to break away from the shackles of the two-stream structure and unify the learning of spatial features and motion information. 
However, the main problem is that this approach simply adds spatial features to motion encodings, so it lacks the mutual enhancement within spatiotemporal features. 
TEA \cite{li2020tea} extends the structure of STM and avoids the extra costs encountered in a two-stream structure. 
However, TEA and the STM share a common problem of using only adjacent temporal features for determining differences to portray the motion features of videos; this technique is prone to pixel-level difference polarization and interframe similarity information loss.
Different from these methods, our proposed CMEM solves the above problems through joint learning and the interaction of spatial and temporal motion features.

For long-range temporal aggregation at large scales, the existing methods usually adopt 3D CNNs. 
A 3D CNN expands the dimensions of convolution kernels, and the new dimensions correspond to the temporal dimensions of videos. 
The original 3D CNN \cite{ji20123d} was first proposed in 2012 and was used to apply a network to the video behavior classification task. 
Then, Tran \cite{tran2015learning} improved upon the 3D CNN structure and proposed the convolutional 3D (C3D) network, which achieved unified spatiotemporal feature modeling by using a 3D CNN on complete video frames.
However, although a 3D CNN is faster than two-stream networks due to the avoidance of additional optical flow inputs, the increase in the dimensionality of the convolution kernels inevitably brings additional computational costs. 
For this reason, some researchers have proposed methods for decomposing the 3D convolution operation into 2D spatial convolution and 1D temporal convolution (also known as (2+1) D convolution) \cite{lin2019temporal, qiu2019learning, diba2017temporal, he2019stnet, tran2019video}. 
Some researchers have also explored the use of 2D CNNs to replace some convolution layers in 3D CNNs; this is called a mixture of 2D CNN and 3D CNN \cite{tran2018closer, xie2018rethinking} models. 
However, after a large number of local convolution operations, the interactions between long-range frames become extremely limited, and it becomes difficult to capture effective spatiotemporal information.

For the above reason, Diba \cite{diba2017temporal} and He \cite{he2019stnet} proposed the temporal 3D ConvNet (T3D) method and spatial temporal network (StNet) respectively.
These two methods model the long-range spatiotemporal relationships in videos through dense connections and self-attention mechanisms respectively.
However, such methods often need to introduce additional parameters or time-consuming operations to improve the interaction of long-range frames.
Our proposed CLIM increases the interaction frequencies of long-range spatiotemporal features through a cascaded residual network structure, combined with an adaptive shift module, which finally achieves long range temporal aggregation at large scales.

\section{Methodology}

\begin{figure}[htbp]
\centering 
\centerline{\includegraphics[width = \linewidth]{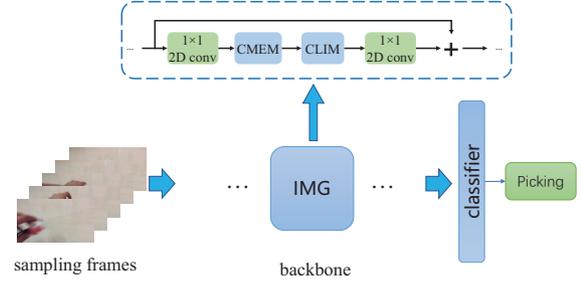}}
\caption{The overall framework of the proposed method.}
\label{fig2}
\end{figure}

The framework of IMG is illustrated in Fig.~\ref{fig2}. 
It is composed of a sampling process, backbone network and classifier. 
First, a frame sequence with a certain length is extracted from the original video through data preprocessing. 
Then, the spatial features and temporal motion features of the video are extracted through the backbone IMG network, including the CMEM and CLIM. 
Finally, the obtained features are passed through the classifier to receive the classification result of the behavior.

\subsection{CMEM}

Motion features directly reflect the behaviors of humans in videos. 
In previous work, behavior recognition approaches based on optical flows only focused on temporal motion patterns at the pixel-level and the learning processes for motion features and spatial features were separated. 
In contrast, our proposed CMEM can recognize motion patterns at the feature-level and unify the learning of motion and spatial features. 
Furthermore, the CMEM also introduces channel attention to suppress the background information as irrelevant noise,
and temporal motion information is enhanced as task-relevant information,
which corresponds to video motion patterns.

\begin{figure}[htbp]
\centering 
\centerline{\includegraphics[width=0.6\linewidth]{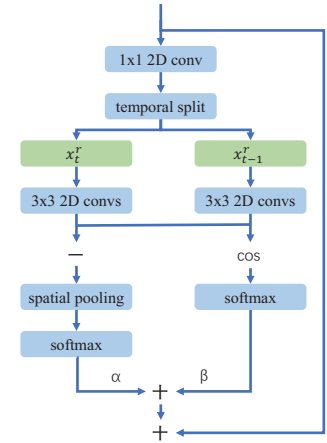}}
\caption{The architecture of the CMEM.}
\label{fig3}
\end{figure}

The architecture of the CMEM is shown in Fig.~\ref{fig3}. 
The input spatiotemporal feature can be expressed as $X[N,T,C,H,W]$, where $N$ is the batch size and $T$ and $C$ are the length of the frame sequence and the number of feature channels, respectively. $H$ and $W$ correspond to the height and width of the features, respectively.
Given an input feature $X$, the number of the feature channels is reduced by a $1 \times 1$ 2D convolution layer to improve the efficiency of the model. 
Then, we obtain $X^r$, which is calculated as:

\begin{equation}
\boldsymbol{X}^{r}=\operatorname{conv}_{\text {prev }} * \boldsymbol{X}, \quad \boldsymbol{X}^{r} \in \mathbb{R}^{N \times T \times C / r \times H \times W}\label{eq1},
\end{equation}

\noindent where $conv_{prev}$ is a $1 \times 1$ 2D convolution layer, $\ast$ indicates the convolution operation, and r is the reduction ratio.
For $X^r$, the spatiotemporal features of the adjacent frames are expressed as $X^r_{(t-1)}$ and $X^r_{(t)}$:

\begin{equation}
\begin{array}{lr}
\boldsymbol{X}_{(t-1)}^{r}=\boldsymbol{X}[N, t, C, H, W], \quad 1 \leq t \leq T, \\ \\
\boldsymbol{X}_{(t)}^{r}=\boldsymbol{X}[N, t, C, H, W], \quad 0 \leq t \leq T-1.

\label{eq2}
\end{array}
\end{equation}

$X^r_{(t-1)}$ and $X^r_{(t)}$ are passed through a $3 \times 3$ convolution layer to prevent excessive mismatches between the two feature vectors. 


Unlike the STM\cite{jiang2019stm} and TEA\cite{li2020tea}, the CMEM collaboratively considers the differences and the cosine similarities between adjacent frames for feature-level motion representation. While a difference tends to polarize the variances of some pixel points between adjacent frames, cosine similarity measures the overall similarity between adjacent frames. This makes the training process smoother. The above process can be expressed as Eqs. (\ref{eq3}) and (\ref{eq4}).

\begin{equation}
\resizebox{.8\hsize}{!}{
$\boldsymbol{M}(t)=\operatorname{conv}_{\text {trans }} * \boldsymbol{X}_{(t)}^{r}-\text { conv }_{\text {trans }} * \boldsymbol{X}_{(t-1)}^{r} ,\quad 1 \leq t \leq T$
}
\label{eq3}
\end{equation}

\begin{equation}
\resizebox{.8\hsize}{!}{
$\boldsymbol{P}(t)=\cos \left(\operatorname{conv}_{\text {trans }} * \boldsymbol{X}_{(t)}^{r}, \quad \operatorname{conv}_{\text {trans }} * \boldsymbol{X}_{(t-1)}^{r}\right), \quad 1 \leq t \leq T$
}
\label{eq4}
\end{equation}

Here, $conv_{trans}$ is a 2D $3 \times 3$ convolution layer, $cos(A, B)$ denotes the cosine similarity between A and B, and the formula is shown in Eq. (\ref{eq5}). $\mathrm{M}(\mathrm{t}) \in \mathrm{R}^{\mathrm{N} \times \mathrm{C} / \mathrm{r} \times \mathrm{H} \times \mathrm{W}}$ and $P(t) \in R^{N \times \mathrm{C} / \mathrm{r} \times 1 \times 1}$ denote the two types of different feature-level motion information. Then, we concatenate $M(1), M(2),...,M(T)$ and $P(1),P(2),...,P(T)$ to obtain the motion feature matrices $\boldsymbol{M}=[M(1),M(2),...,M(T)]$ and $\boldsymbol{P}=[P(1),P(2),...,P(T)]$. To ensure that the temporal dimensions of the matrices are the same as those of the input vectors, we fill the ends of the two matrices with 0s.

\begin{equation}
\cos (A, B)=\frac{A \cdot B}{\|A\|\,\|B\|} 
\label{eq5}
\end{equation}

After that, the spatial details of the feature maps are less important since the purpose of the CMEM is to enhance the motion-sensitive channels in spatiotemporal features. Hence, the dimensions of $\boldsymbol{M}$ are then changed to $\mathrm{N} \times \mathrm{T} \times \mathrm{C} / \mathrm{r} \times 1 \times 1$ through average pooling:

\begin{equation}
\boldsymbol{M}^{q}=\operatorname{Pool}(\boldsymbol{M}), \quad \boldsymbol{M}^{s} \in \mathbb{R}^{N \times T \times C / r \times 1 \times 1}\label{eq6}
\end{equation}

Both $\boldsymbol{M}^q$ and $\boldsymbol{P}$ reflect the differences between adjacent frames, and they are called video motion features. In addition, $\boldsymbol{M}^q$ and $\boldsymbol{P}$ are complementary, and we can then obtain the complete motion matrix F by spatial weighting in the CMEM:

\begin{equation}
\boldsymbol{F}=\alpha * \boldsymbol{M}^{q}+\beta * \boldsymbol{P}, \quad \boldsymbol{F} \in \mathbb{R}^{N \times T \times C / r \times 1 \times 1},\label{eq7}
\end{equation}

\noindent where $\alpha$ and $\beta$ are hyperparameters that denote the contributions of the motion features to the final attention weights. To enhance the motion-sensitive information of the spatiotemporal features, F is passed through a $1 \times 1$ 2D convolution layer that expands its number of channel to the same as that of the input X. Finally, the motion matrix is normalized by a sigmoid function and a linear transformation to obtain the final attention weights $\boldsymbol{F}^s$:

\begin{equation}
\boldsymbol{F}^{\boldsymbol{s}}=2 * \delta\left(\operatorname{conv}_{\operatorname{exp} } * \boldsymbol{F}-1\right), \quad \boldsymbol{F} \in \mathbb{R}^{N \times T \times C \times 1 \times 1},\label{eq8}
\end{equation}

\noindent where $Conv_{exp}$ denotes a $1 \times 1$ convolution layer, $\ast$ indicates the convolution operation, and $\delta$ denotes the sigmoid function. Then, the attention weights and the input spatiotemporal features are multiplied in the channel dimension to achieve the discovery and enhancement of motion-sensitive information in the spatiotemporal features. In addition, a residual layer is applied to ensure that the background information is not lost during the training process. After introducing the residual structure, the calculation of the input features can be expressed as:

\begin{equation}
\boldsymbol{X}^{o}=\boldsymbol{X}+\boldsymbol{X} \odot \boldsymbol{F}^{\boldsymbol{s}}, \quad \boldsymbol{X}^{o} \in \mathbb{R}^{N \times T \times C \times H \times W}
\label{eq9}
\end{equation}

\noindent where $\boldsymbol{X}^{o}$ is the final output and $\odot$ denotes the channel-wise multiplication operation.

\subsection{CLIM}

In this section, we first propose an adaptive shift module to reconstruct frame sequences, as this module provides the basis for more flexibility and reliability among the long-range interactions. Then, this module is utilized to achieve long-term motion feature integration. 

\subsubsection{Adaptive shift Module}

Lin \cite{lin2019temporal} proposed a temporal shift module (TSM) via the creative use of the shift operator; this module can realize the rearrangement of frame sequences. Given an input $X[N,T,C,H,W]$, the frame shift module in the TSM can be expressed as:

\begin{equation}
\begin{array}{lr}
\boldsymbol{X}[N, t, C, H, W]=\boldsymbol{X}[N, t+1, C, H, W] \\
\quad 1 \leq t \leq T-1, \quad 1 \leq c \leq C / 8,\quad \text { left shift }
\label{eq10}
\end{array}
\end{equation}

\begin{equation}
\begin{array}{lr}
\boldsymbol{X}[N, t, C, H, W]=\boldsymbol{X}[N, t-1, C, H, W]\\
\quad 2 \leq t \leq T, \quad C / 8<c \leq C / 4,\quad \text { right shift }\label{eq11}
\end{array}
\end{equation}


\begin{equation}
\begin{array}{lr}
\boldsymbol{X}[N, t, C, H, W]=\boldsymbol{X}[N, t, C, H, W]\\
\quad 1 \leq t \leq T, \quad C / 4<c \leq C,\quad \text { unchanged }\label{eq12}
\end{array}
\end{equation}





\noindent where $N$ is the batch size, $T$ is the length of the frame sequence, and $C$ is the number of feature channels. $H$ and $W$ correspond to the height and width of the feature, respectively.

The process of frame shifting is equivalent to the following transformation: fixing the dimensions of the $N$, $H$ and $W$ of the original features, and replacing some features in the $T$ and $C$ dimensions \cite{li2020tea}. This process is exactly equivalent to the convolution process expressed by Eq. (\ref{eq13}).

\begin{equation}\label{eq13}
\begin{array}{lll}
\mathbf{X}_{\mathrm{rp}}=\text { Reshape }(\mathbf{X}), & \mathbf{X} \in \mathbb{R}^{N \times T \times C \times H \times W}, & \mathbf{X}_{\mathrm{rp}} \in \mathbb{R}^{N H W \times C \times T} \\
\mathbf{X}_{\mathrm{ss}}=\mathbf{K} * \mathbf{X}_{\mathrm{rp}}, & \mathbf{X}_{\mathrm{ss}} \in \mathbb{R}^{N H W \times C \times T}, & \mathbf{K} \in \mathbb{R}^{C \times 1 \times 3} \\
\mathbf{K}[\mathrm{c}, 1, k]=1, & 1 \leq \mathrm{c} \leq C / 8, & k=3 \\
\mathbf{K}[\mathrm{c}, 1, k]=0, & 1 \leq \mathrm{c} \leq C / 8, & k=1,2 \\
\mathbf{K}[\mathrm{c}, 1, k]=1, & C / 8<\mathrm{c} \leq C / 4, & k=1 \\
\mathbf{K}[\mathrm{c}, 1, k]=0, & C / 8<\mathrm{c} \leq C / 4, & k=2,3 \\
\mathbf{K}[\mathrm{c}, 1, k]=1, & C / 4<\mathrm{c} \leq C, & k=2 \\
\mathbf{K}[\mathrm{c}, 1, k]=0, & C / 4<\mathrm{c} \leq C, & k=1,3
\end{array}
\end{equation}

The dimentions of the input are modified by $Reshape(X)$; after transformation, we obtain $X_{rp} \in R^{NHW\times C\times T}$. $X_{ss} $is the result of convolution, $\ast$ indicates the convolution operation, $K$ denotes the convolution kernel, and $K[c,1,k]$ denotes the point in the convolution kernel with coordinates $(c,1,k)$.

\begin{figure}[htbp]
\centering 
\centerline{\includegraphics[width=0.4\linewidth]{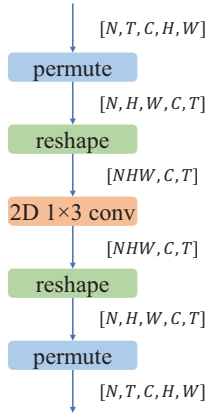}}
\caption{The structure of the adaptive shift module.}
\label{fig4}
\end{figure}

Therefore, the frame shifting operation in the TSM \cite{lin2019temporal} can be regarded as a special convolution transformation. Based on this, we propose an adaptive shift module, and the details of the module are shown in Fig.~\ref{fig4}, where permute denotes vector rearrangement and reshape denotes vector transformation.

Compared with the fixed frame sequence reconstruction process in the TSM \cite{lin2019temporal}, the process in the adaptive shift module no longer relies on hyperparameters and empirical values. This module can be easily integrated into a deep network and participate in the backpropagation process, providing a basis for more flexible and reliable spatiotemporal interactions among long-range frames.

\subsubsection{CLIM}

Previous behavior recognition methods \cite{tran2015learning, sun2015human} usually stacked local temporal convolutions to achieve long-range temporal modeling. However, after a large number of local convolutions, the interactions between long-range frames become limited and it is difficult to capture effective spatiotemporal information. Therefore, the CLIM is proposed to achieve more effective long-term motion information modeling, and this module is inspired by Res2Net \cite{gao2019res2net}.

\begin{figure}[htbp]
\centering 
\centerline{\includegraphics[width=0.3\textwidth]{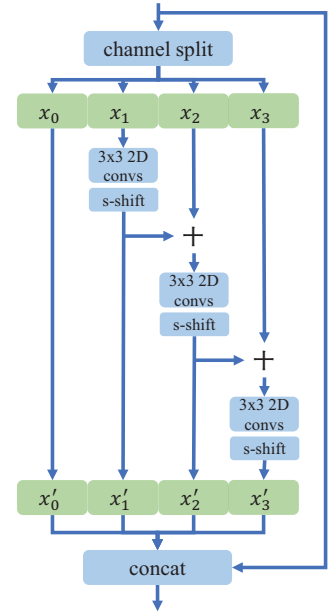}}
\caption{The structure of the proposed CLIM.}
\label{fig5}
\end{figure}

The architecture of the CLIM is shown in Fig.~\ref{fig5}. The CLIM divides the input spatiotemporal features into several subsets based on their channels. For each subset, the features are further extracted by a set of local convolutions, including one adaptive shift layer and a $3 \times 3$ 2D convolution layer. We use the residual structures between the adjacent subsets of the CLIM, and these residual blocks not only allow the CLIM to form various receptive fields but also increase the frequency of spatiotemporal feature interactions since the adaptive shift layer makes the frame sequence change through local convolution. At the same time, spatiotemporal interactions do not lead to redundant parameters and time-consuming operations. Finally, the integration of long-term motion features is completed by concatenating the features extracted from each subset.

Given an input feature $X[N,T,C,H,W]$, a typical processing method is to extract the temporal and spatial information \cite{lin2019temporal, qiu2019learning, diba2017temporal, he2019stnet, tran2019video} from this feature via a local temporal convolution and a spatial convolution; this process is also called (2+1)D convolution. Based on the convolution and the Res2Net architecture, we propose a new spatiotemporal feature interaction method. First, $X$ is divided into four equal slices $X0,X1,X2,X3$ in the channel dimension, and the shape of each is $[N,T,C/4,H,W]$. Except for the first slice, the remaining three slices are passed through a convolution group consisting of 2D CNN and adaptive shift modules.

The slices pass through the convolution group and then interact with each other through residual blocks of spatiotemporal features, thus forming a cascaded residual structure. This process can be expressed as Eq. (\ref{eq14}).

\begin{equation}
\begin{aligned}
\boldsymbol{X}_{i}^{\prime} &=\boldsymbol{X}_{i}, & i=0 \\
\boldsymbol{X}_{i}^{\prime} &=\operatorname{SSM}\left(\operatorname{conv}_{s p t} * \boldsymbol{X}_{i}\right), & i=1 \\
\boldsymbol{X}_{i}^{\prime} &=\operatorname{SSM}\left(\operatorname{conv}_{s p t} *\left(\boldsymbol{X}_{i}+\boldsymbol{X}_{i-1}^{\prime}\right)\right), & i=2,3
\end{aligned}
\label{eq14}
\end{equation}

\noindent where $SSM(X)$ denotes the process conducted by the adaptive shift layer on features, $conv_{spt}$ denotes a $3 \times 3$ 2D convolution layer, $\ast$ denotes the convolution operation, $X_i$ denotes the slice of the $i_{th}$ input, and $X^{'}_{i}$ denotes the output corresponding to the $i_{th}$ slice $(X^{'}\in R^{N\times T\times C/4\times H\times W},i=0,1,2,3)$.

The cascade structure of the CLIM provides multiple scales of receptive fields in the feature extraction process. In the CLIM, the receptive fields of different slices are enlarged to different degrees through residual blocks. For example, the receptive field of the second slice is twice as large as that of the first slice, and the perceptual field of the fourth slice is four times as large as that of the first slice. Furthermore, due to the continuous reconstruction of frame sequences by the adaptive shift module, the CLIM effectively enhances the probability of spatiotemporal interaction for long-range frames.

\subsection{IMG Motion Feature Module}
Finally, the CMEM and CLIM are concatenated with ResNet blocks to establish our IMG motion feature module. The approach is illustrated in the backbone of Fig.~\ref{fig2}. To unify the dimensions, $1 \times 1$ convolutions are added to the head and tail of our IMG module. Then, we stack the IMG module to construct the whole behavior recognition network.

In addition, the cross-entropy loss is utilized as our loss function. It can be expressed as:

\begin{equation}
L=\frac{1}{N} \sum_{i} L_{i}=-\frac{1}{N} \sum_{i} \sum_{c=1}^{M} y_{i c} \log \left(p_{i c}\right)
\end{equation}

\noindent where M denotes the number of classes, N denotes the number of samples, $y_{i c}$ equals 1 if the label of the $i_{th}$ sample is $c$, and $p_{i c}$ denotes the predicted probability of class $c$.

\section{Experiments}

\subsection{Datasets and Evaluation Metrics}

Our proposed approach is evaluated on three datasets: HMDB51 \cite{kuehne2011hmdb}, UCF101 \cite{soomro2012ucf101} and Something-Something \cite{goyal2017something}. Among them, HMDB51 can be divided into 51 classes with a total of 6849 video clips, and each class has at least 101 samples. UCF101 can be divided into 101 classes with a total of 13320 video samples. Compared with the previous two datasets, Something-Something has larger numbers of samples and behavioral classes, with a total of 108,499 videos and 174 classes of behaviors.

HMDB51 and UCF101 are commonly used in behavior recognition. Something-Something contains more training data, and its videos are clearer; therefore, we train our model on this dataset. The other two datasets are fine-tuned on the model pretrained on Something-Something.

Accuracy is an important metric for measuring the results obtained on a multiclass classification task such as behavior recognition. For a multiclass classification task, if the total number of samples is $N$, the total number of classes is $M$, and $r_i$ samples are correctly predicted for class i in the predicted results, the accuracy can be described as:

\begin{equation}
\operatorname{Accuracy}=\frac{\sum_{i=1}^{M} r_{i}}{N} \quad(i=1,2, \ldots, M)\label{eq16}
\end{equation}

More categories are contained in Something-Something, and some classes are highly similar. Therefore, our model uses both top-1 accuracy and top-5 accuracy as evaluation metrics.

\subsection{Implementation Details}

Previous work \cite{lin2019temporal, qiu2019learning, diba2017temporal, he2019stnet, tran2019video} has demonstrated that pretrained models can significantly improve the training efficiency and reliability of the developed models. Therefore, we apply the publicly released Res2Net50 model to initialize the parameters.

In addition, although Res2Net performs better than ResNet on various image-based tasks, \cite{li2020tea} suggested that no significant improvement is achieved for video behavior recognition tasks. Thus, Res2Net is utilized as our backbone.

Two types of model training methods are available. The first is similar to the TSM \cite{lin2019temporal}, STM \cite{jiang2019stm}, TEA \cite{li2020tea}, etc. All batch normalization layers in the model are enabled during training. For the Something-Something dataset, the batch size, initial learning rate, weight decay and dropout rate are set to 32, 0.01, 1e4 and 0.5, respectively. The model training process adopts a stochastic gradient descent strategy. The number of training epochs is 50, and the learning rate is decreased by a factor of 10 at 30, 40, and 45 epochs.

Another strategy is aimed at UCF101 and HMDB51, where the training process is based on the Something-Something pretrained model. Therefore, the hyperparameters are somewhat distinct from those of the previous training method. To maintain the reliability of training, the parameters of each batch normalization layer are frozen except for those of the first layer. In this mode, the batch size, initial learning rate, weight decay and dropout rate are set to 32, 0.001, 5e4 and 0.8, respectively. The learning rate is decreased by a factor of 10 at 10 and 20
epochs. The training process stops after 25 epochs.

\subsection{Experimental Results}

\subsubsection{Comparisons with the State-of-the-Art Methods}
\begin{table*}[hbt]
\begin{center}
\caption{Comparison among the results obtained by IMG and other state-of-the-art methods on Something-Something.}
\resizebox{\textwidth}{!}{
\begin{tabular}{ccccccc}
\hline
\textbf{Method} & \textbf{Backbone} & \textbf{FLOPs} & \textbf{Frames} $\times$ \textbf{Crops} $\times$ \textbf{Clips} & \textbf{Pre-train} & \textbf{Val Top-1($\%$)} & \textbf{Val Top-5($\%$)} \\
\hline
\text { I3D-RGB }\cite{carreira2017quo} & \text { 3D ResNet50 } & 153 \text{G} $\times$ 3 $\times$ 2 & 32 $\times$ 3 $\times$ 2 & \multirow{3}*{ImageNet+Kinetics400} & 41.6 & 72.2 \\
\text { NL I3D-RGB }\cite{wang2018videos} & \text { 3D ResNet50 } & 168 \text{G} $\times$ 3 $\times$ 2 & 32 $\times$ 3 $\times$ 2 & ~ & 44.4 & 76.0 \\
\text { NL I3D+GCN-RGB }\cite{wang2018videos} & \text { 3D ResNet50+GCN } & 303 \text{G} $\times$ 3 $\times$ 2 & 32 $\times$ 3 $\times$ 2 & ~ & 46.1 & 46.8 \\
\hline
\text { STM-RGB }\cite{jiang2019stm} & \text { ResNet50 } & 33 \text{G} $\times$ 3 $\times$ 10 & 8 $\times$ 3 $\times$ 10 & \text { ImageNet } & 49.2 & 79.3 \\
\hline
\text { TSM-RGB }\cite{lin2019temporal} & \multirow{3}*{ResNet50}  & 33 \text{G} $\times$ 1 $\times$ 1 & 8 $\times$ 1 $\times$ 1 &  \multirow{3}*{ImageNet+Kinetics400} & 43.4 & 47.3 \\
\text { TSM-RGB }\cite{lin2019temporal} & ~ & 65 \text{G} $\times$ 1 $\times$ 1 & 16 $\times$ 1 $\times$ 1 & ~ & 44.8 & 74.5 \\
\text { TSM-(RGB+FLOW) }\cite{lin2019temporal} & ~ & \text{N} / \text{A} & 16 $\times$ 1 $\times$ 1+16 $\times$ 1 $\times$ 1 & ~ & 50.2 & 79.5 \\
\hline
\text{TEA}\cite{gao2019res2net} & \multirow{2}*{ Res2Net50 }
 & 35 \text{G} $\times$ 1 $\times$ 1 & 8 $\times$ 1 $\times$ 1 & \multirow{2}*{ ImageNet } & 48.9 & 78.1 \\
\text{TEA}\cite{gao2019res2net} & ~ & 35 \text{G} $\times$ 3 $\times$ 10 & 8 $\times$ 3 $\times$ 10 & ~ & 51.7 & 80.5 \\

\hline
\hline
CMEM    &  \multirow{2}*{Res2Net50}  & 34G$\times$ 1$\times$ 1  & 8 $\times$ 1 $\times$ 1 & \multirow{2}*{ImageNet} & \CMEMsomet & \CMEMsomeT \\
CMEM    &  ~  & 34G$\times$ 3$\times$ 10  & 8 $\times$ 3 $\times$ 10 & ~ & \CMEMsomett & \CMEMsomeTT \\
\hline
IMG    &  \multirow{2}*{Res2Net50}  & 37G$\times$ 1$\times$ 1  & 8 $\times$ 1 $\times$ 1 & \multirow{2}*{ImageNet} & \textbf{\CMEMCLIMsomet} & \textbf{\CMEMCLIMsomeT} \\
IMG    &  ~  & 37G$\times$ 3$\times$ 10  & 8 $\times$ 3 $\times$ 10 & ~ & \textbf{\CMEMCLIMsomett} & \textbf{\CMEMCLIMsomeTT} \\
\hline
\end{tabular}
}
\label{tab_all}
\end{center}
\end{table*}

To analyze the effectiveness of our proposed model and to demonstrate that the CLIM has an enhancing effect on the CMEM, the accuracy of our proposed model is compared with that of the current state-of-the-art model on Something-Something. The comparison domains include the backbones, floating point operations(FLOPs), data processing strategies and pretrained models. To ensure the credibility of the comparison experiments, the model results are validated by using 2 different data processing strategies, and the main purpose is to balance the speed and accuracy of the model. One is an efficient strategy that uses one central crop for each video frame. The other is a precise strategy in which 10 different clips are randomly sampled from the video, and the final score is calculated by averaging all clips' scores. The results are shown in Table~\ref{tab_all}.

Compared with these models, IMG achieves state-of-the-art performance in terms of top-1 accuracy and top-5 accuracy under the same strategy. When the efficient strategy is used to process the data, the top-1 accuracy and top-5 accuracy are \CMEMCLIMsomet\% and \CMEMCLIMsomeT\%, respectively, and when the exact strategy is used, the top-1 accuracy and top-5 accuracy reach \CMEMCLIMsomett\% and \CMEMCLIMsomeTT\%. The FLOPs exhibit a slight increase in our proposed method compared to those produced by TEA \cite{li2020tea}, but our performance is increased significantly (\CMEMCLIMsomet\% $vs.$ 48.9\%).

We model the short-distance motions between adjacent frames and the large-scale associations of long time sequences with our proposed CMEM and CLIM. It can be concluded that the CMEM mainly learns motion patterns from adjacent frames and carries out the interactions between spatiotemporal features. As the network deepens and the receptive field of the network increases, the CMEM can learn motion information under a wider range of videos. However, its modeling of motion features in a long temporal field still has some room to improve, which means that the CMEM lacks the ability to discover the large-scale associations of long time sequences. 

Based on this, the CLIM further improves the effectiveness of the CMEM with a cascaded residual network structure and adaptive shift, thus greatly increasing the interaction frequencies of the long-range frames in the model and enhancing the model's ability to discover the large-scale associations of long time sequences.

Finally, to validate our method on more benchmark datasets, comparison results obtained on UCF101 and HMDB51 are reported in Table~\ref{tab5}.

\begin{table}[htbp]
\begin{center}
\caption{Comparison among the results obtained by IMG and other state-of-the-art methods on HMDB51 and UCF101.}
\begin{tabular}{cccc}
\hline
\textbf{Method} & \textbf{Backbone} & \makecell[c]{\textbf{HMDB51} \\ \textbf{Top-1(\%)}} & \makecell[c]{\textbf{UCF101} \\ \textbf{Top-1(\%)}}\\
\hline
TSM\cite{lin2019temporal}     & ResNet50     & 70.7  & 94.5 \\
STM\cite{jiang2019stm}    & ResNet50    & 72.2  & 96.2\\
TEA\cite{li2020tea}    & Res2Net50    & 73.3  & 96.9\\
\hline
\hline
IMG   & Res2Net50    &  \textbf{\CMEMCLIMhmdbt}  & {\CMEMCLIMucft}\\
\end{tabular}
\label{tab5}
\end{center}
\end{table}

The results show that our proposed IMG achieves \CMEMCLIMhmdbt\%~on HMDB51, outperforming other methods. Nevertheless, it can be observed that the result on UCF101 is slightly decreased. From our analysis, this is because some behavioral categories in UCF101 have almost no long-time associations (e.g., playing the guitar or flute). Thus, the introduction of the CLIM instead increases the complexity of the model and reduces its robustness.

In summary, our proposed IMG behavior recognition network achieves effective modeling of the spatial and motion information in videos while avoiding excessive parameter scales and improves the overall model's behavior recognition accuracy to achieve the current state-of-the-art performance.

\subsubsection{Ablation Study}

To ensure the effectiveness of our proposed CMEM module and its significant ability to model spatiotemporal motion features, we design an ablation experiment to confirm the performance differences yielded by the CMEM module; this experiment is conducted on different backbone networks by comparing the behavior recognition performances of different combinations of the 2D CNN and the CMEM. The results are shown in Table~\ref{tab1}.

\begin{table}[htbp]
\begin{center}
\caption{Ablation results.}
\begin{tabular}{ccc}
\hline
\textbf{Method}& \textbf{Val Top-1(\%)} & \textbf{Val Top-5(\%)}\\
\hline
(2+1) D ResNet     & 45.9  & 75.2 \\
(2+1) D Res2Net    & 46.2  & 75.6\\
\hline
\hline
CMEM + ResNet    & 48.7  & 78.1\\
CMEM + Res2Net    &  \textbf{\CMEMsomet}  & \textbf{\CMEMsomeT}\\
\end{tabular}
\label{tab1}
\end{center}
\end{table}

\begin{figure}[htbp]
\centering 
\centerline{\includegraphics[width=0.5\textwidth]{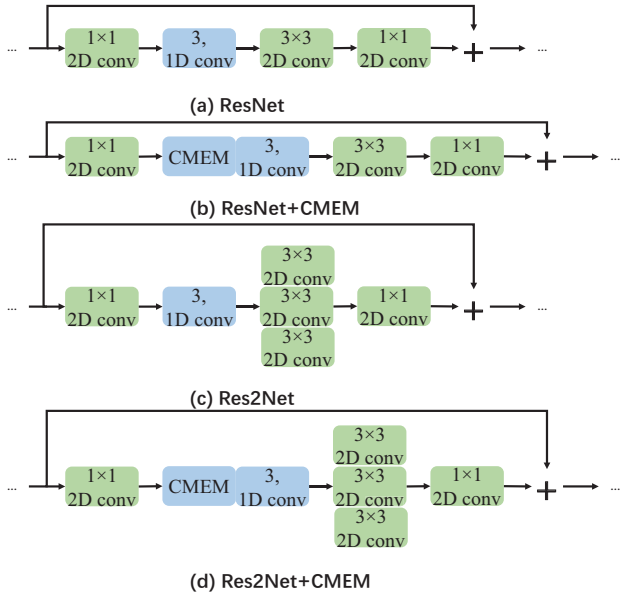}}
\caption{Four types of combinations are applied in the ablation experiment.}
\label{fig6}
\end{figure}

Four types of combinations are tested in the experiment, namely, the ResNet residual block, ResNet+CMEM residual block, Res2Net residual block, and Res2Net+CMEM residual block. The structure of each residual block is shown in Fig.~\ref{fig6}.

From the results, we can conclude that our proposed CMEM can effectively improve the behavior recognition accuracy of the model. Res2Net is more complex than ResNet, yet the corresponding behavior recognition performance is not significantly improved. In contrast, the performance is greatly improved by introducing the CMEM in both backbone networks, and the CMEM+Res2Net achieves the best behavior recognition results, with top-1 and top-5 accuracies that reach \CMEMsomet\%~ and \CMEMsomeT\%~, respectively. Thus, it can be concluded that our proposed CMEM can achieve significantly improved the behavior recognition performance over that of the base ResNet and Res2Net residual structures.

Although we use Res2Net as the backbone network, it does not clearly improve the behavior recognition accuracy over that achieved with ResNet. By comparing the results in the first two rows or the last two rows of Table~\ref{tab1}, it is obvious that although Res2Net achieves much higher performance than ResNet in many image processing tasks such as image classification and image object detection \cite{gao2019res2net}, the difference between the backbone networks does not significantly affect the accuracy of behavior recognition. Thus, we can conclude that the performance improvement is due to the spatiotemporal motion modeling capability of the CMEM, which excites the motion-sensitive part of the original features and improves the model's ability to represent short-range motion information.

\subsubsection{Evaluation of the adaptive Shift Module}

The adaptive shift module is equivalent to the frame reconstruction process in the TSM under specific cases. To verify this fact to fully utilize the performance of the adaptive shift module, we design different CLIM convolution sets (2D CNN+shift) as a way to verify the effectiveness of the adaptive shift module. The results are shown in Table~\ref{tab_CLIM}.

\begin{table}[h]
\begin{center}
\caption{Evaluation of the adaptive shift module.}
\begin{tabular}{cccc}
\hline
\textbf{Method}& \textbf{Backbone}& \textbf{Val Top-1(\%)} & \textbf{Val Top-5(\%)}\\
\hline
TSM-RGB & ResNet50 & 43.4 & 73.2 \\
\hline
\hline
Structure(a) & Res2Net50     & \at  & \aT \\
Structure(b) & Res2Net50     & \bt  & \bT \\
Structure(c) & Res2Net50     & \ct  & \cT \\
Structure(d) & Res2Net50     & \textbf{\CLIMsomet} & \textbf{\CLIMsomeT} \\
\end{tabular}
\label{tab_CLIM}
\end{center}
\end{table}

\begin{figure}[htbp]
\centering 
\centerline{\includegraphics[width=0.5\textwidth]{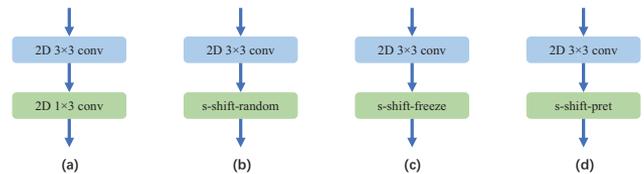}}
\caption{Four CLIM convolution groups. }
\label{fig7}
\end{figure}

The table includes the backbone network used by each method and the corresponding accuracy metrics, which indicate the different CLIM convolution groups. The structure of each convolution group is shown in Fig.~\ref{fig7}. $s\raisebox{0mm}{-}shift\raisebox{0mm}{-}random$ indicates that the parameters of the convolution layer of the adaptive shift module are initialized randomly, $s\raisebox{0mm}{-}shift\raisebox{0mm}{-}freeze$ indicates that its parameters are frozen, and $s\raisebox{0mm}{-}shift\raisebox{0mm}{-}pret$ indicates that its parameters are initialized to the values in Eq. \eqref{eq13}.

\begin{figure}[htbp]
\centering 
\centerline{\includegraphics[width=0.5\textwidth]{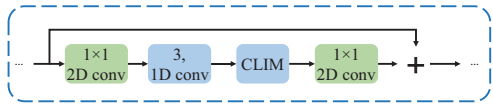}}
\caption{The backbone utilized in this experiment.}
\label{fig8}
\end{figure}

To avoid the influence of the CMEM module on the experimental results, we use the backbone shown in Fig.~\ref{fig8} for this experiment.

The results show that the adaptive shift module can effectively improve the accuracy of the behavior recognition method (up to \CLIMsomet\% top-1 accuracy and \CLIMsomeT\% top-5 accuracy on Something-Something), which positively reflects the superior ability of the CLIM in terms of modeling long-term motion features.

By comparing the results of structure (a) and structure (b) (\at\% $vs.$ \bt\% for top-1 accuracy and \aT\% $vs.$ \bT\% for top-5 accuracy), we find that the results are very similar, which supports the fact that the adaptive shift module with randomly initialized parameters is similar to a 1D convolution in the temporal dimension, meaning that they have close semantic representations.

From the results of structure (b) and structure (c) (\bt\% $vs.$ \ct\% for top-1 accuracy and \bT\% $vs.$ \cT\% for top-5 accuracy), it can be found that the adaptive shift module based on the frame sequence reconstruction strategy in the STM exhibits better performance than the adaptive shift module with randomly initialized parameters. One possible explanation for this phenomenon is that the STM's frame sequence reconstruction strategy is a local optimum obtained after a number of different strategy choices. In contrast, the adaptive shift module with randomly initialized parameters tends to have difficulty converging.

To take full advantage of the local optimal results of the STM, the parameters of $s\raisebox{0mm}{-}shift\raisebox{0mm}{-}freeze$ are used for the initialization of structure (d). The adaptive shift module achieves the best results on Something-Something (\CLIMsomet\% accuracy for top-1 accurate and \CLIMsomeT\% accuracy for top-5 accurate). Thus, the CLIM can achieve more frequent long-distance frame interactions and improves the model's ability to model long-term motion information.

In summary, our proposed adaptive shift module is close to the 1D temporal convolution layer in the (2+1)D convolution at the semantic level, which can effectively extract of temporal information in features. Through a specific initialization of its parameter settings, the adaptive shift module can achieve a further performance improvement.

\section{Conclusion}

In this paper, we propose a behavior recognition method based on the integration of multigranular motion features that is composed of a CMEM and a CLIM. Specifically, the CMEM achieves short-range motion encoding between adjacent frames through simultaneous learning of spatial and temporal motion features. The CLIM enables more frequent long-range frame interactions, thus making up for the shortcomings of the CMEM. The two proposed methods are then concatenated to achieve joint spatial and temporal motion feature learning in both short-range and long-range situations.


\end{document}